\documentclass{article}

\usepackage{arxiv}

\usepackage[utf8]{inputenc} 
\usepackage[T1]{fontenc}    
\usepackage{hyperref}       
\usepackage{url}            
\usepackage{booktabs}       
\usepackage{amsfonts}       
\usepackage{nicefrac}       
\usepackage{microtype}      
\usepackage{lipsum}
\usepackage{graphicx}
\graphicspath{ {./images/} }

\usepackage{multirow}
\usepackage[table,xcdraw]{xcolor}
\usepackage{booktabs}
\usepackage{array}
\newcolumntype{P}[1]{>{\centering\arraybackslash}m{#1}}
\usepackage{wrapfig}

\usepackage{subcaption}

\usepackage{listings}
\title{LLMs Still Can't Plan; Can LRMs?\\A Preliminary Evaluation of OpenAI's o1 on PlanBench}

\author{%
  Karthik Valmeekam\thanks{Equal Contribution} \\
   SCAI, Arizona State University\\
   \texttt{kvalmeek@asu.edu} \\
   \And
  Kaya Stechly$^*$ \\
  SCAI, Arizona State University\\
  \texttt{kstechl@asu.edu} \\
   \And
  Subbarao Kambhampati \\
  SCAI, Arizona State University \\
   \texttt{rao@asu.edu}
}

\begin{document}
\maketitle
\begin{abstract}
The ability to plan a course of action that achieves a desired state of affairs has long been considered a core competence of intelligent agents and has been an integral part of AI research since its inception. With the advent of large language models (LLMs), there has been considerable interest in the question of whether or not they possess such planning abilities. PlanBench \cite{valmeekam2024planbench}, an extensible benchmark we developed in 2022, soon after the release of GPT3, has remained an important tool for evaluating the planning abilities of LLMs. Despite the slew of new private and open source LLMs since GPT3, progress on this benchmark has been surprisingly slow. OpenAI claims that their recent o1 (Strawberry) model has been specifically constructed and trained to escape the normal limitations of autoregressive LLMs--making it a new kind of model: a Large Reasoning Model (LRM). Using this development as a catalyst, this paper takes a comprehensive look at how well current LLMs and new LRMs do on PlanBench. As we shall see, while o1's performance is a quantum improvement on the benchmark, outpacing the competition, it is still far from saturating it. This improvement also brings to the fore questions about accuracy, efficiency, and guarantees which must be considered before deploying such systems. 
\end{abstract}
\setcounter{footnote}{0}
\section{Introduction}
The recent release of OpenAI's o1 (Strawberry)~\cite{o1systemcard} brings with it the opportunity to both freshly evaluate progress on PlanBench and to consider directions for extending the benchmark. In particular, unlike the LLMs which came before it, which can roughly be viewed as approximate retrievers, o1 seems to have been trained to be an approximate reasoner.\footnote{We speculate that the complete system learns to improve its ability to make appropriate Chain-of-Thought (CoT) moves useful for reasoning in a pretraining RL step with synthetic data, and does inference time prompt-specific rollouts \cite{rao-tweet}. In other words, it may be an RL-trained system in the same vein as AlphaGo, but where the `moves' being generated and evaluated are Chains of Thought.} Following OpenAI\footnote{Per the blogpost announcing the model: "A new series of reasoning models [...] for complex reasoning tasks this is a significant advancement and represents a new level of AI capability. Given this, we are resetting the counter back to 1 and naming this series OpenAI o1."\cite{openai-o1-intro}}, we draw a distinction between previous Large Language Models and o1, a Large Reasoning Model (or LRM), as its new (unknown) architecture, operation, and capabilities all seem to be fundamentally different from those of vanilla LLMs, both at pretraining phase and at inference time. To properly evaluate this new kind of model and understand its abilities and limitations will require new tools and evaluation methods, especially if details of the overall model structure are kept secret and internal traces remain inaccessible to outside researchers.\footnote{There are reports that OpenAI is threatening to revoke access to o1 from anyone who tries to extract internal reasoning traces~\cite{edwards2024ban}.}

Since PlanBench~\cite{valmeekam2024planbench} first debuted on arXiv in 2022, we have continually retested new models on one particular subset: a static dataset of 600 three to five block problems. Even though LLMs have gotten ever larger and required substantially more investment per model, their performance on even the simplest planning problems has never come close to saturating this test set, and the improvements we have seen have not been robust or generalizable~\cite{stechly2024chain}. Our benchmark has thus served as a useful marker of LLM progress (or lack thereof) on planning and reasoning tasks, though it is important to note that this analysis--especially when confined to a static test set--can only serve as an upper bound on performance. When we first designed PlanBench, it was intended to be more than just this set, but rather to be an extensible suite of tools for evaluating LLM planning capabilities. Now that LRMs score so highly on at least parts of the original test set, those tools will become ever more important for future evaluations.

In this preliminary evaluation (at time of writing, o1-preview and o1-mini have only been out a week, and the full o1 model is yet to be released), we examine the performance jump that these new Large Reasoning Models promise. We record the slow progress we've observed in vanilla LLM performance since the release of the benchmark, discuss o1's performance, and then tackle the question of how PlanBench's domains and tests can be elaborated on in order to remain relevant metrics for LRMs. We argue that, to be complete, new approaches to measuring LRM reasoning capabilities must take into account efficiency, cost, and guarantees.
\section{State-of-the-Art LLMs Still Can't Plan}
\begin{table}
\centering
\small
\begin{tabular}{@{}P{0.8cm}P{0.6cm}P{1.2cm}P{1cm}P{1.1cm}P{1cm}P{1cm}P{1cm}P{1cm}P{1cm}P{1cm}P{1cm}@{}}
\toprule
\multirow{2}{*}{\textbf{Domain}}                                                           & \multirow{2}{*}{\textbf{Shots}}                     & \multicolumn{2}{c}{\textbf{Claude Models}}                                                                                                  & \multicolumn{4}{c}{\textbf{OpenAI GPT-4 Models}}                                                                                                                                                                                                             & \multicolumn{2}{c}{\textbf{LLaMA Models}}                                                                                          & \multicolumn{2}{c}{\textbf{Gemini Models}}                                                                                            \\ \cmidrule(l){3-12} 
       &                                                     & \textbf{\begin{tabular}[c]{@{}c@{}}Claude 3.5\\ (Sonnet)\end{tabular}} & \textbf{\begin{tabular}[c]{@{}c@{}}Claude 3\\ (Opus)\end{tabular}} & \textbf{GPT-4o}                                             & \textbf{\begin{tabular}[c]{@{}c@{}}GPT-4o\\ -mini\end{tabular}} & \textbf{GPT-4}                                              & \textbf{\begin{tabular}[c]{@{}c@{}}GPT-4\\ Turbo\end{tabular}} & \textbf{\begin{tabular}[c]{@{}c@{}}LLaMA\\ 3.1 405B\end{tabular}} & \textbf{\begin{tabular}[c]{@{}c@{}}LLaMA\\ 3 70B\end{tabular}} & \textbf{\begin{tabular}[c]{@{}c@{}}Gemini \\ 1.5 Pro\end{tabular}} & \textbf{\begin{tabular}[c]{@{}c@{}}Gemini \\ 1 Pro\end{tabular}} \\ \midrule
\multirow{2}{*}{\textbf{\begin{tabular}[c]{@{}l@{}}Blocks\\ world\end{tabular}}}           & \begin{tabular}[c]{@{}l@{}}One\\ Shot\end{tabular}  & \begin{tabular}[c]{@{}c@{}}\textbf{346/600}\\ \textbf{(57.6\%)}\end{tabular}             & \begin{tabular}[c]{@{}c@{}}289/600 \\ (48.1\%)\end{tabular}        & \begin{tabular}[c]{@{}c@{}}170/600\\ (28.3\%)\end{tabular}  & \begin{tabular}[c]{@{}c@{}}49/600\\ (8.1\%)\end{tabular}        & \begin{tabular}[c]{@{}c@{}}206/600 \\ (34.3\%)\end{tabular} & \begin{tabular}[c]{@{}c@{}}138/600\\ (23\%)\end{tabular}       & \begin{tabular}[c]{@{}c@{}}284/600\\ (47.3\%)\end{tabular}        & \begin{tabular}[c]{@{}c@{}}76/600\\ (12.6\%)\end{tabular}      & \begin{tabular}[c]{@{}c@{}}101/600\\ (16.8\%)\end{tabular}         & \begin{tabular}[c]{@{}c@{}}68/600\\ (11.3\%)\end{tabular}        \\ \cmidrule(l){2-12} 
       & \begin{tabular}[c]{@{}l@{}}Zero\\ Shot\end{tabular} & \begin{tabular}[c]{@{}c@{}}329/600\\ (54.8\%)\end{tabular}             & \begin{tabular}[c]{@{}c@{}}356/600 \\ (59.3\%)\end{tabular}        & \begin{tabular}[c]{@{}c@{}}213/600 \\ (35.5\%)\end{tabular} & \begin{tabular}[c]{@{}c@{}}53/600\\ (8.8\%)\end{tabular}        & \begin{tabular}[c]{@{}c@{}}210/600 \\ (34.6\%)\end{tabular} & \begin{tabular}[c]{@{}c@{}}241/600\\ (40.1\%)\end{tabular}     & \begin{tabular}[c]{@{}c@{}}\textbf{376/600}\\ \textbf{(62.6\%)}\end{tabular}        & \begin{tabular}[c]{@{}c@{}}205/600\\ (34.16\%)\end{tabular}    & \begin{tabular}[c]{@{}c@{}}143/600\\ (23.8\%)\end{tabular}         & \begin{tabular}[c]{@{}c@{}}3/600\\ (0.5\%)\end{tabular}          \\ \midrule
\multirow{2}{*}{\textbf{\begin{tabular}[c]{@{}l@{}}Mystery\\ Blocks\\ world\end{tabular}}} & \begin{tabular}[c]{@{}l@{}}One\\ Shot\end{tabular}  & \begin{tabular}[c]{@{}c@{}}19/600\\ (3.1\%)\end{tabular}               & \begin{tabular}[c]{@{}c@{}}8/600 \\ (1.3\%)\end{tabular}           & \begin{tabular}[c]{@{}c@{}}5/600\\ (0.83\%)\end{tabular}    & \begin{tabular}[c]{@{}c@{}}0/600\\ (0\%)\end{tabular}           & \begin{tabular}[c]{@{}c@{}}\textbf{26/600} \\ \textbf{(4.3\%)}\end{tabular}   & \begin{tabular}[c]{@{}c@{}}5/600\\ (0.83\%)\end{tabular}       & \begin{tabular}[c]{@{}c@{}}21/600\\ (3.5\%)\end{tabular}          & \begin{tabular}[c]{@{}c@{}}15/600\\ (2.5\%)\end{tabular}       & -                                                                  & \begin{tabular}[c]{@{}c@{}}2/500 \\ (0.4\%)\end{tabular}         \\ \cmidrule(l){2-12} 
       & \begin{tabular}[c]{@{}l@{}}Zero\\ Shot\end{tabular} & \begin{tabular}[c]{@{}c@{}}0/600\\ (0\%)\end{tabular}                  & \begin{tabular}[c]{@{}c@{}}0/600\\ (0\%)\end{tabular}              & \begin{tabular}[c]{@{}c@{}}0/600\\ (0\%)\end{tabular}       & \begin{tabular}[c]{@{}c@{}}0/600\\ (0\%)\end{tabular}           & \begin{tabular}[c]{@{}c@{}}1/600 \\ (0.16\%)\end{tabular}   & \begin{tabular}[c]{@{}c@{}}1/600\\ (0.16\%)\end{tabular}       & \begin{tabular}[c]{@{}c@{}}\textbf{5/600}\\ \textbf{(0.8\%)}\end{tabular}           & \begin{tabular}[c]{@{}c@{}}0/600\\ (0\%)\end{tabular}          & -                                                                  & \begin{tabular}[c]{@{}c@{}}0/500\\ (0\%)\end{tabular}            \\ \bottomrule
\end{tabular}
\vspace{0.2cm}
\caption{Performance on 600 instances from the Blocksworld and Mystery Blocksworld domains across large language models from different families, using both zero-shot and one-shot prompts. Best-in-class accuracies are bolded.}
\label{tab:llm_results}
\end{table}
PlanBench remains a challenging benchmark for vanilla LLMs (massive transformer models which have been fine-tuned via RLHF), and their lackluster performance on even our easiest test set leads us to continue to believe that planning cannot be generally and robustly solved by approximate retrieval alone. In Table~\ref{tab:llm_results}, we present the results of running current and previous generation LLMs on a static test set of 600 three to five block Blocksworld problems, as well as on a set of 600 semantically identical but syntactically obfuscated instances which we call Mystery Blocksworld. Across these models, the best performance on regular Blocksworld is achieved by LLaMA 3.1 405B with $62.6\%$ accuracy. Despite the underlying problems being identical, Mystery Blocksworld performance lags far behind--no LLM achieves even 5\% on our test set--and performance on one version of the domain does not clearly predict performance on the other.\footnote{We do not provide Mystery Blocksworld data for Gemini 1.5 Pro only because we haven't been able to generate it. The model refuses to produce any output, instead claiming that responding to these queries would be harmful. We include this output in Appendix~\ref{a:gemini_harm}.}

In the original paper, we tested both natural language prompts and PDDL, and found that vanilla language models perform better when tested on the former, even though natural language prompts can introduce uncertainty due to polysemanticity and syntactic ambiguity. To make our comparisons "fair" for the models being tested, the results we have been reporting are the higher accuracy natural language prompting numbers.

LLMs are highly capable of providing translations between equivalent representations~\cite{olmo2021gpt3}. This fact, combined with their significantly higher performance on the unobfuscated version of the Blocksworld domain, predicts that--if they are capable of composing reasoning operations--the performance gap between Mystery Blocksworld and classic Blocksworld should shrink substantially if the translation from Mystery Blocksworld back into Blocksworld is explicitly provided. However, when we provide this in the prompt (see Appendix~\ref{a:mystery_mapping}), performance only improves a very small amount: GPT-4 achieves 10\%.

We also find that, contrary to previous claims, one-shot prompting is \textit{not} a strict improvement over zero-shot. In fact, for many models it seems to do significantly worse!\footnote{While the reverse is generally true for Mystery Blocksworld problems, it's important to note that the performance of vanilla LLMs on Mystery Blocksworld has consistently and uniformly been poor (the same as it was when this benchmark was first released), so those results do not provide too clear a picture. Most models do not solve even a single instance in zero-shot mode, and only one (LLaMA 3.1 405B) manages more than one.} This is most notable in our tests of LLaMA family models.

The original iteration of our benchmark did not take efficiency into consideration, as the time taken by a vanilla LLM to produce some output is only dependent on the length of that output, but otherwise independent of the semantic content or difficulty of the instance. However, as LRMs adaptively vary their time taken and dollar cost per instance in response to the input, measuring efficiency has become much more important. As a comparison point between LRMs and LLMs, we compute prompting costs across models and present them in Table~\ref{tab:costs}.
\section{From Approximate Retrieval to Approximate Reasoning: Evaluating o1}
\begin{figure}
\centering
\begin{subfigure}{.5\textwidth}
  \centering
  \includegraphics[width=1\linewidth]{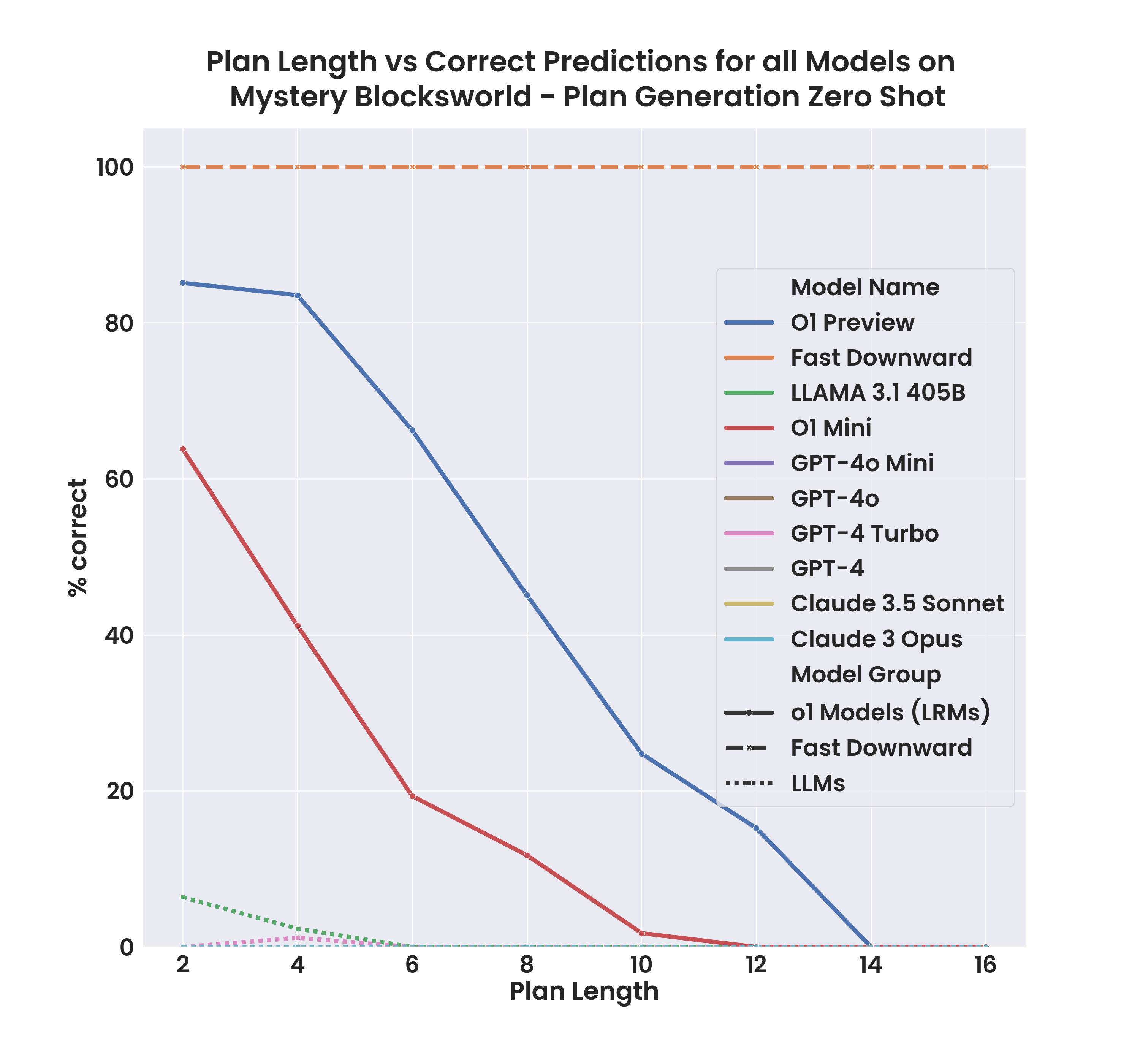}
  \label{fig:length_zero_shot}
\end{subfigure}%
\begin{subfigure}{.5\textwidth}
  \centering
  \includegraphics[width=1\linewidth]{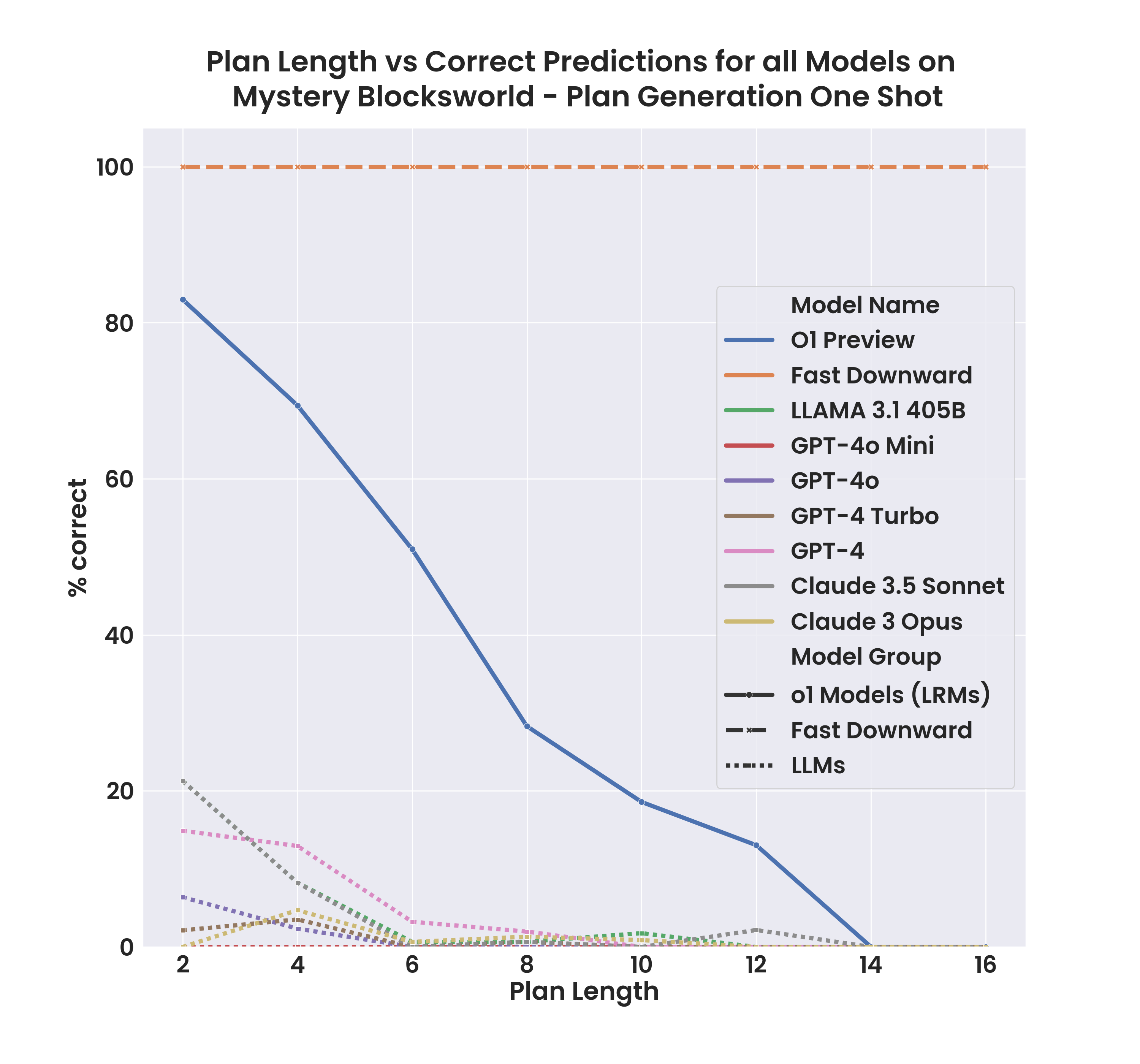}
  \label{fig:length_one_shot}
\end{subfigure}
\caption{These examples are on Mystery Blocksworld. Fast Downward, a domain-independent planner~\cite{helmert2006fast} solves all given instances near-instantly with guaranteed perfect accuracy. LLMs struggle on even the smallest instances. The two LRMs we tested, o1-preview and o1-mini, are surprisingly effective, but this performance is still not robust, and degrades quickly with length.}
\label{fig:length}
\end{figure}
\begin{figure}
\centering
\begin{subfigure}{.45\textwidth}
  \centering
  \includegraphics[width=1\linewidth]{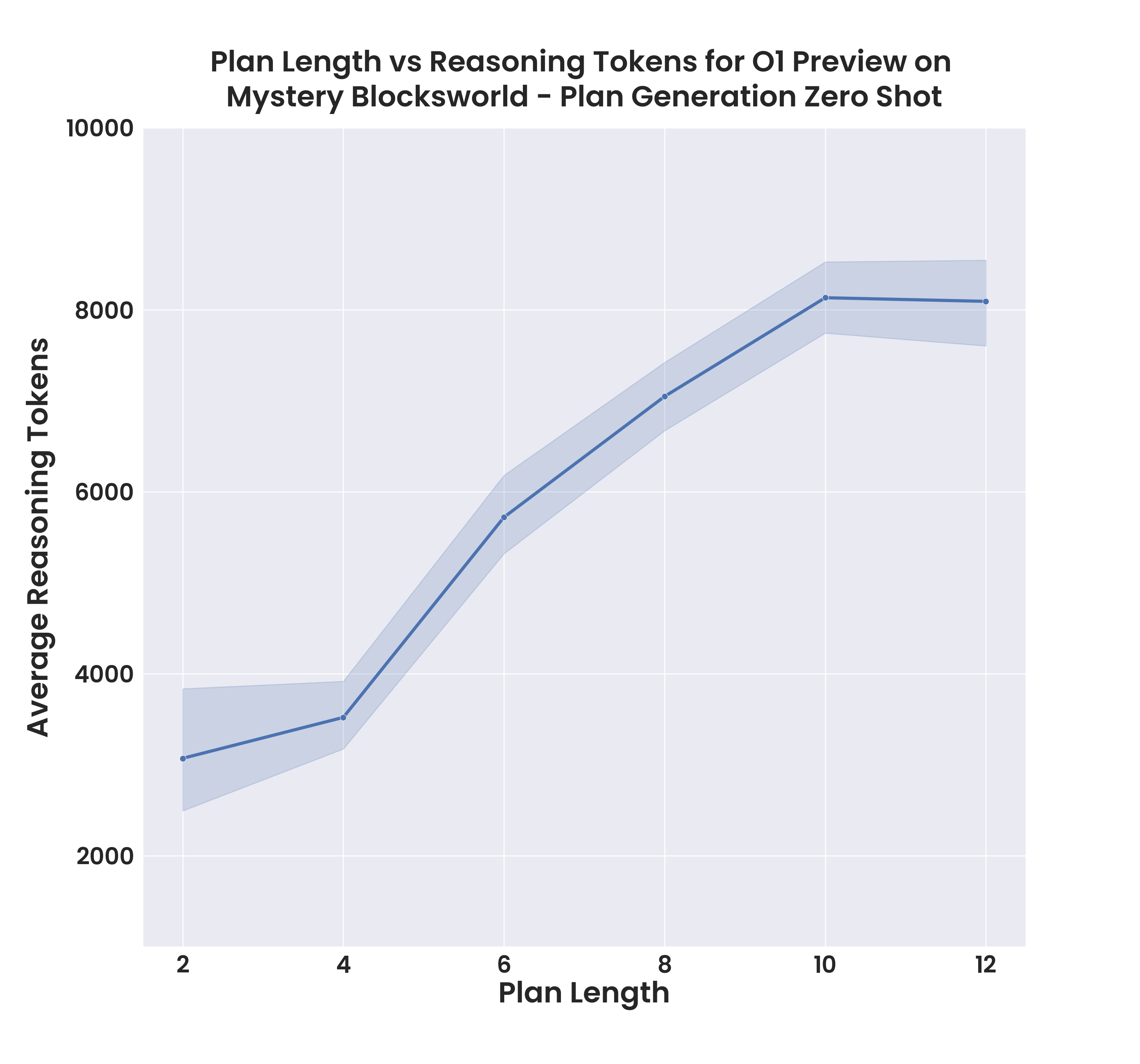}
  \label{fig:reasoning_tokens_mb}
  \caption{Average number of reasoning tokens used by o1-preview when solving Mystery blocksworld instances correlates with the length of the optimal plan required to solve the instance.}
\end{subfigure}%
\quad
\begin{subfigure}{.45\textwidth}
  \centering
  \includegraphics[width=1\linewidth]{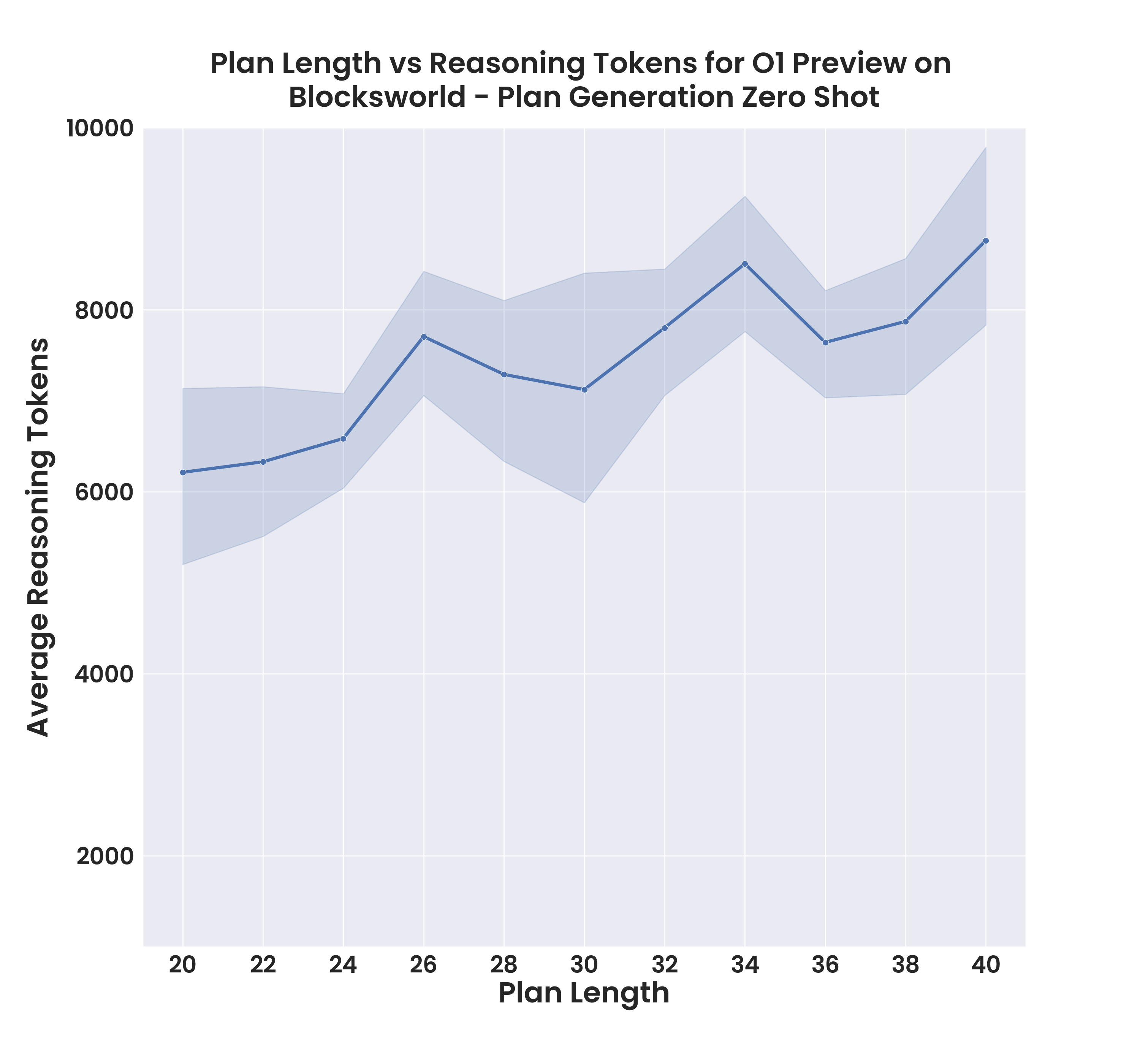}
  \label{fig:reasoning_tokens_bw}
  \caption{Average number of reasoning tokens used by o1-preview does not increase as expected with the number of steps required to solve the instance.}
\end{subfigure}
\caption{}
\label{fig:reasoning_tokens}
\end{figure}
\begin{table}[]
\centering
\small
\begin{tabular}{@{}llcccccc@{}}
\toprule
\multirow{2}{*}{\textbf{Domain}}                                                                  & \multirow{2}{*}{\textbf{Shots}} & \multicolumn{3}{c}{\textbf{Instances correct}}                                                                                                                                       & \multicolumn{3}{c}{\textbf{\begin{tabular}[c]{@{}c@{}}Average Time Taken (in secs)\end{tabular}}} \\ \cmidrule(l){3-8} 
    &                                 & \textbf{o1-preview}                                        & \textbf{o1-mini}                                            & \textbf{Fast Downward}                                    & \textbf{o1-preview}             & \textbf{o1-mini}             & \textbf{Fast Downward}             \\ \midrule
\textbf{Blocksworld}                                                                              & Zero Shot                       & \begin{tabular}[c]{@{}c@{}}587/600\\ (97.8\%)\end{tabular} & \begin{tabular}[c]{@{}c@{}}340/600 \\ (56.6\%)\end{tabular} & \begin{tabular}[c]{@{}c@{}}600/600\\ (100\%)\end{tabular} & 40.43                           & 10.84                        & 0.265                              \\ \midrule
\multirow{2}{*}{\textbf{\begin{tabular}[c]{@{}l@{}}Mystery\\ Blocksworld\end{tabular}}} & One Shot                        & \begin{tabular}[c]{@{}c@{}}247/600\\ (41.6\%)\end{tabular} & -                                                           & \begin{tabular}[c]{@{}c@{}}600/600\\ (100\%)\end{tabular} & 82.03                           & -                            & 0.265                              \\ \cmidrule(l){2-8} 
                  & Zero Shot                       & \begin{tabular}[c]{@{}c@{}}317/600\\ (52.8\%)\end{tabular} & \begin{tabular}[c]{@{}c@{}}115/600\\ (19.1\%)\end{tabular}  & \begin{tabular}[c]{@{}c@{}}600/600\\ (100\%)\end{tabular} & 83.37                           & 35.54                        & 0.265                              \\ \midrule
\textbf{\begin{tabular}[c]{@{}l@{}}Randomized Mystery \\ Blocksworld\end{tabular}}                & Zero Shot                       & \begin{tabular}[c]{@{}c@{}}224/600\\ (37.3\%)\end{tabular} & -                                                           & \begin{tabular}[c]{@{}c@{}}600/600\\ (100\%)\end{tabular} & 111.11                          & -                            & 0.265                              \\ \bottomrule
\end{tabular}
\vspace{0.2cm}
\caption{Performance and average time taken on 600 instances from the Blocksworld, Mystery Blocksworld and Randomized Mystery Blocksworld domains by OpenAI's o1 family of large reasoning models and Fast Downward}
\label{o1_results}
\end{table}

Many researchers, including us, have argued that "standard" autoregressive LLMs generate outputs via approximate retrieval, and that, while they show impressive performance on a range of System 1 tasks, they are unlikely to achieve more System 2-like approximate reasoning capabilities critical for planning tasks (c.f. \cite{rao-nyas}). Until now, the best way to coax sound planning capabilities out of LLMs has been to pair them with external verifiers in a generate-test framework, in what are called LLM-Modulo systems \cite{kambhampati2024llms,trinh2024solving}. o1 attempts to supplement an underlying LLM with System 2-like abilities in a different way. 

As far as we can tell, o1 combines an underlying LLM, most likely a modified GPT-4o, into an RL-trained system that steers the creation, curation, and final selection of private Chain-of-Thought reasoning traces. Exact details are currently sparse, and so we can only speculate about its exact mechanisms. Our best guess is that there are two major differences between o1 and LLMs: an additional reinforcement learning pre-training phase (perhaps to learn the q-values of different CoTs from massive amounts of synthetic data) and a new adaptively scaling inference procedure (maybe it further refines learned q-values by something like rollout before selecting a particular CoT; see \cite{rao-tweet}). Regardless, what looks clear from the detail available is that this model is fundamentally different in nature from previous LLMs.

\paragraph{Evaluating LRMs on the Original Test Set:} We test o1-preview and o1-mini on the static PlanBench test set.\footnote{While for previous models, we had the model itself enforce the plan format we wanted, some modifications had to be made to accurately test o1's abilities. In its current form, o1-preview does not always conform to explicit formatting restrictions. This is right in line with OpenAI's injunction to keep o1 prompts "simple and direct" \cite{openai-reasoning-guide2024}. In order to extract the generated plans, we used GPT-4o-mini to translate them into PDDL, and wrote a small Python parser to strip any remaining extraneous symbols before evaluating each proposed plan.} The full results can be seen in Table~\ref{o1_results}. Our 600 Blocksworld instances range from three to five blocks, and require plans of between 2 to 16 steps to solve. Far surpassing any LLM, o1 correctly answers 97.8\% of these instances. On Mystery Blocksworld, the model does not maintain this level of performance, but it does far surpass all previous models, answering 52.8\% correctly. To test whether the exact obfuscation might be compromised because of data contamination, we also generated a new obfuscation using completely random strings, and presented these problems in a new, semantically equivalent prompt format with fully specified and unambiguous PDDL descriptions of both the domain and problem. This is presented in the table as Randomized Mystery Blocksworld. While performance did dip further, 37.3\% of instances are answered correctly, sharply contrasting the flat zeroes we saw when testing previous models.

\begin{figure}
    \centering
    \includegraphics[width=\linewidth]{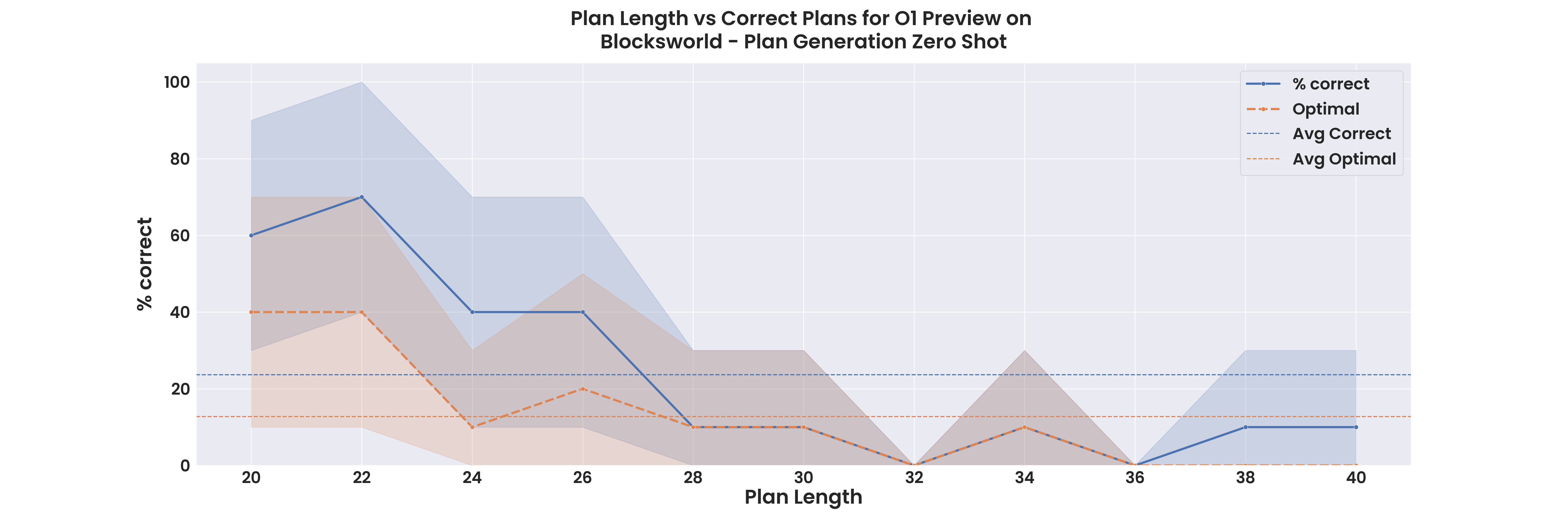}
    \caption{Extending even the (regular, not obfuscated) Blocksworld dataset to problems requiring greater numbers of steps worsens the performance of o1-preview. When tested on 110 instances which each require at least 20 steps to solve, it only manages 23.63\%.}
    \label{fig:obsr_o1}
\end{figure}
\paragraph{Accuracy with Increased Problem Size:} Standard LLM chain-of-thought prompting approaches are brittle, do not robustly scale with problem size, and fail to induce general algorithmic procedure-following~\cite{stechly2024chain}. We test the models on a set of 110 larger Blocksworld problems (see Figure \ref{fig:obsr_o1}). Problems in this set range from 6 to 20 blocks in length and require 20 to 40 step optimal plans. Without any obfuscation, we see performance quickly degrade from the 97.8\% reported earlier. In fact, over these 110 instances, o1-preview only manages 23.63\%, and most of this accuracy comes from correctly solving problems which require fewer than 28 steps. While these models are overall impressive, this shows that their performance is still far from robust.
\begin{table}
    \centering
\small
\begin{tabular}{@{}llcc@{}}
\toprule
\multirow{2}{*}{\textbf{Domain}}                                                  & \multicolumn{1}{c}{\multirow{2}{*}{\textbf{Shots}}} & \multicolumn{2}{c}{\textbf{o1-preview}}             \\ \cmidrule(l){3-4} 
                                                                                  & \multicolumn{1}{c}{}                                & \textbf{True Negatives} & \textbf{False Negatives} \\ \midrule
\textbf{Blocksworld}                                                              & 0-Shot                                              & 27\%                     & 0\%                      \\ \midrule
\textbf{\begin{tabular}[c]{@{}l@{}}Randomized Mystery\\ Blocksworld\end{tabular}} & 1-Shot                                              & 16\%                     & 11.5\%                   \\ \bottomrule
\end{tabular}
\vspace{0.2cm}
\caption{Rate of claiming that a problem is impossible by OpenAI's o1-preview on 100 unsolvable and 600 solvable instances in the Blocksworld and Randomized Mystery Blocksworld domains. The True Negative rate is the percent of unsolvable instances that were \textit{correctly} marked as unsolvable. The False Negative rate is the percent of solvable instances that were \textit{incorrectly} marked as unsolvable. Previous models are not shown in this table as their true negative and false negative rates were generally 0\% across the board.}
\label{o1_unsolvable}
\end{table}

\paragraph{Performance on Unsolvable Instances:} While planning problems normally require the agent to formulate a course of action to achieve a goal, an equally valid use of planning abilities is to recognize that a given goal \textit{cannot} be accomplished by any plan. A real-world example of this is network vulnerability analysis, where an agent may wish to certify that no plan of attack exists for a specified system~\cite{boddy-aaai-cybersecurity}. So far, LLMs have struggled to recognize that some problems cannot be solved, instead confidently confabulating nonsensical answers. o1 was launched with the claim that it has started to overcome this issue, and can now accurately identify unsolvable problems~\cite{noam-unsolvable-tweet}. To test this systematically, we modified 100 instances from our original three to five block test set by adding one \verb|on(x,y)|-type conjunct to each instance's goal state, making the goal unsatisfiable.\footnote{We ensured our instances were unambiguous by giving the full PDDL representation of both the domain and the instance, to avoid quibbles such as "A is on B because A is on C and C is on B" where the model redefines the meanings of an ambiguous natural language statement.} The results are in Table~\ref{o1_unsolvable}. On Blocksworld, only 27\% of all instances were correctly and explicitly identified by o1 as unsolvable. In 19\% of all cases, the model returned a dot or some kind of "[empty plan]" marker, without any explanation or indication of unsolvability. We consider these incorrect, as "empty plan" is only the correct answer if the goal is already satisfied. In the remaining 54\% of cases, the model generated a full (and therefore impossible and incorrect!) plan.

On Randomized Mystery Blocksworld, these numbers are worse: 16\% of cases were correctly identified as unsolvable, 5\% returned an empty plan, and the remaining 79\% were answered with a full (impossible or goal-unsatisfying) plan. Therefore, unsolvable instances continue to be a problem for LRMs. Furthermore, this ability to sometimes note impossible plans correctly comes at a cost: now the model sometimes falsely claims that solvable problems are actually unsolvable. On Randomized Mystery Blocksworld, 11.5\% of instances are incorrectly claimed to be impossible. These results can be seen in Table~\ref{o1_unsolvable}.
\begin{table}[]
\centering
\small
\begin{tabular}{@{}P{1.2cm}P{1cm}P{1.1cm}P{1cm}P{1cm}P{1cm}P{1cm}P{1cm}|cc@{}}
\toprule
\multicolumn{10}{c}{\textbf{Costs per 100 instances (in USD)}}                                                                                                                                                                                                                                                                                                                                                                                                                                                  \\ \midrule
\multicolumn{8}{c}{\textbf{Large Language Models}}                                                                                                                                                                                                                                                                                                                                                                                                        & \multicolumn{2}{c}{\textbf{Large Reasoning Models}} \\ \midrule
\textbf{\begin{tabular}[c]{@{}c@{}}Claude 3.5\\ (Sonnet)\end{tabular}} & \textbf{\begin{tabular}[c]{@{}c@{}}Claude 3\\ (Opus)\end{tabular}} & \textbf{GPT-4o} & \textbf{\begin{tabular}[c]{@{}c@{}}GPT-4o\\ -mini\end{tabular}} & \textbf{GPT-4} & \textbf{\begin{tabular}[c]{@{}c@{}}GPT-4\\ Turbo\end{tabular}} & \textbf{\begin{tabular}[c]{@{}c@{}}Gemini \\ 1.5 Pro\end{tabular}} & \textbf{\begin{tabular}[c]{@{}c@{}}Gemini \\ 1 Pro\end{tabular}} & \textbf{o1-preview}        & \textbf{o1-mini}       \\ \midrule
\$0.44                                                                 & \$1.70                                                             & \$0.65          & \$0.02                                                          & \$1.80         & \$1.20                                                         & \$0.33                                                             & \$0.03                                                           & \$42.12                    & \$3.69               \\ \bottomrule
\end{tabular}
\vspace{0.2cm}
\caption{Cost per 100 instances (in USD). LRMs are significantly more expensive than LLMs.}
\label{tab:costs}
\end{table}
\paragraph{Accuracy/Cost Tradeoffs and Guarantees:} 
With LRMs showing better performance on planning problems, our evaluations must explicitly take into account the trade-offs that come from choosing general models over established deep and narrow systems. While o1-preview may provide higher accuracy than LLMs, it still fails to provide any correctness guarantees, and it is unclear that it is at all cost-effective. Unlike previous models, whose APIs only charge based on the number of input tokens and the number of output tokens (usually at a rate that is five times higher for the latter), o1's price-per-call includes a surcharge based on the number of "reasoning tokens" it used--tokens generated as part of inference and not revealed to the user--which are charged at the significantly higher output token rate. Currently, end users have no control over the number of these tokens generated, a number which is expanded or limited by the model in its own opaque way. Since the launch of these models less than a week ago, we have already run up a bill of \$1897.55 for just the o1 model experiments on this benchmark!\footnote{The rich irony of researchers using tax payer provided research funds to pay private companies like OpenAI to evaluate their private commercial models is certainly not lost on us.})


The early version of o1-preview that we have access to seems to be limited in the number of reasoning tokens it uses per problem, as can be seen in the leveling off in Figure~\ref{fig:reasoning_tokens} and more clearly from the scatterplot in Appendix~\ref{fig:nodes_tokens}. This may be artificially deflating both the total cost and maximum performance. If the full version of o1 removes this restriction, this might improve overall accuracy, but it could also lead to even less predictable (and ridicuously high!) inference costs. o1-mini is cheaper, but generally less performant.

Without exposing the ability to scale inference time to particular specifications, influence the internal `thinking' process in task-specific ways, or ensure that intermediate steps are evaluated by trusted or sound verifiers, the o1 models are a coarse-grained choice in the space of cost, inference time, guarantees, and performance trade-offs. They aren't, however, the only choices in that space, and reasonable LRM evaluations must take this into account (see similar arguments in \cite{katz-efficiency,sayash-kapoor-agents}). 

Classical planners like Fast Downward~\cite{helmert2006fast} achieve 100\% on our dataset in a fraction of the time, compute, and cost, while providing {\em guarantees that their answers are correct.} Running Fast Downward on a personal computer was essentially free in dollar terms and averaged 0.265 seconds per instance, which is many orders of magnitude faster than the average o1 clock times listed in table~\ref{o1_results}. It is also generally predictable, and can be scaled to harder instances very directly. Vanilla LLMs are typically very good at translating problems between formats, and could be used to do so in concert with a classical planner at a fraction of the cost of LRMs (e.g. \cite{olmo2021gpt3, stone-optimal-llm}). For problems which don't have simple PDDL domain and instance specifications, LLM-Modulo systems may be a safer and cheaper approach: run a smaller, faster LLM in a loop with a sound verifier, so that the combined system will only output guaranteed correct solutions (e.g. \cite{kambhampati2024llms, romera2024mathematical, trinh2024solving}). Previous testing showed that this LLM-Modulo approach can already achieve 82\% accuracy on a subset of our original Blocksworld test set, as well as 70\% in the Logistics domain (see Table 4 in the Results section of \cite{valmeekam2023planning}).

The correctness guarantees provided by these latter two methods are sorely lacking in LRMs like o1. A general reasoning system cannot be deployed in safety critical and non-ergodic domains if it continues to confidently make incorrect plans. o1 is a fully black box system, even more so than previous models, and OpenAI's decision to not only keep the architecture under wraps and hide the reasoning traces, but to warn away and even ban anyone who attempts to understand what is going on inside them~\cite{edwards2024ban}, makes interpretability nearly impossible, and reduces trust in the system overall.\footnote{The current model is also set to a default temperature of 1.0, which further reduces replicability and interpretability--for any given problem, it is never clear whether the result is merely the result of stochasticity. This compounds a problem with OpenAI models that has existed since at least GPT3. Temperature 0 never gave deterministic outputs, and worse, the logprobs provided by the OpenAI API for any given prompt have long been known to fluctuate wildly \cite{xuan-logprobs-tweet}.}

\paragraph{o1's Creative Justifications:} 
While our main focus has been on providing a quantitative evaluation of o1's performance on PlanBench, we have also noticed an o1 idiosyncracy that is worth commenting on. When the model gives an incorrect answer, it also sometimes provides a creative, but nonsensical, justification for its decision. It is almost as if o1 has gone from \textit{hallucinating} to \textit{gaslighting}! In one case, it decided that an unsolvable problem is solvable because a goal condition, while not present in the final state, had been true for at some point during the execution, and thus should continue to count. In another, it declared that \verb|on(a,c)| was true because, as it explained in a brief parenthetical, a was on b which was on c, and was thus a was somewhere above c, which should count as being "on top" of it. As we mentioned earlier, we changed our unsolvable instance prompts from natural language to PDDL in order to make it extremely clear that divergences from our exact definitions were disallowed.

\section{Conclusion}

We took a fresh look at the planning capabilities of both SOTA LLMs, and examined the performance of OpenAI's new o1 models on PlanBench. Over time, LLMs have improved their performance on vanilla Blocksworld--with the best performing model, LlaMA 3.1 405B, reaching 62.5\% accuracy. However, their dismal performance on the obfuscated ("Mystery") versions of the same domain betrays their essentially approximate retrieval nature. In contrast, the new o1 models, which we call LRMs (Large Reasoning Models)--in keeping with OpenAI's own characterizations--not only nearly saturates the original small instance Blockworld test set, but shows the first bit of progress on obfuscated versions. Encouraged by this, we have also evaluated o1's performance on longer problems and unsolvable instances, and found that these accuracy gains are not general or robust. We also discussed the critical accuracy/efficiency tradeoffs that are brought up by the fact that o1 that uses (and charges for) significant inference-time compute, as well as how it compares to other LLM-based approaches (such as LLM-Modulo \cite{kambhampati2024llms}) and dedicated solvers. We hope this research note gives a good snapshot of the planning capabilities of LLMs and LRMs as well as useful suggestions for realistically evaluating them. 

\section*{Acknowledgements}
We acknowledge support and sprited discussions with fellow lab members including Atharwa Gundawar (who, in a parallel study, found that o1 doesn't fare well in scheduling benchmarks like TravelPlan~\cite{xie2024travelplanner} and Natural Plan~\cite{natural-plan}).
This research is supported in part by ONR grant N0001423-1-2409, and gifts from Qualcomm and Amazon.
\bibliographystyle{is-unsrt}  
\bibliography{references} 
\newpage
\appendix
\section*{Appendix}
\renewcommand{\thesubsection}{\Alph{subsection}}
\subsection{Gemini 1.5 Pro Response to Mystery Blocksworld}
\label{a:gemini_harm}
\begin{verbatim}
finish_reason: SAFETY

safety_ratings {
  category: HARM_CATEGORY_SEXUALLY_EXPLICIT
  probability: NEGLIGIBLE
}
safety_ratings {
  category: HARM_CATEGORY_HATE_SPEECH
  probability: NEGLIGIBLE
}
safety_ratings {
  category: HARM_CATEGORY_HARASSMENT
  probability: NEGLIGIBLE
}
safety_ratings {
  category: HARM_CATEGORY_DANGEROUS_CONTENT
  probability: MEDIUM
}
\end{verbatim}

\subsection{o1 Token Use Versus Problem Difficulty}
\begin{figure}[h]
    \centering
    \includegraphics[width=\linewidth]{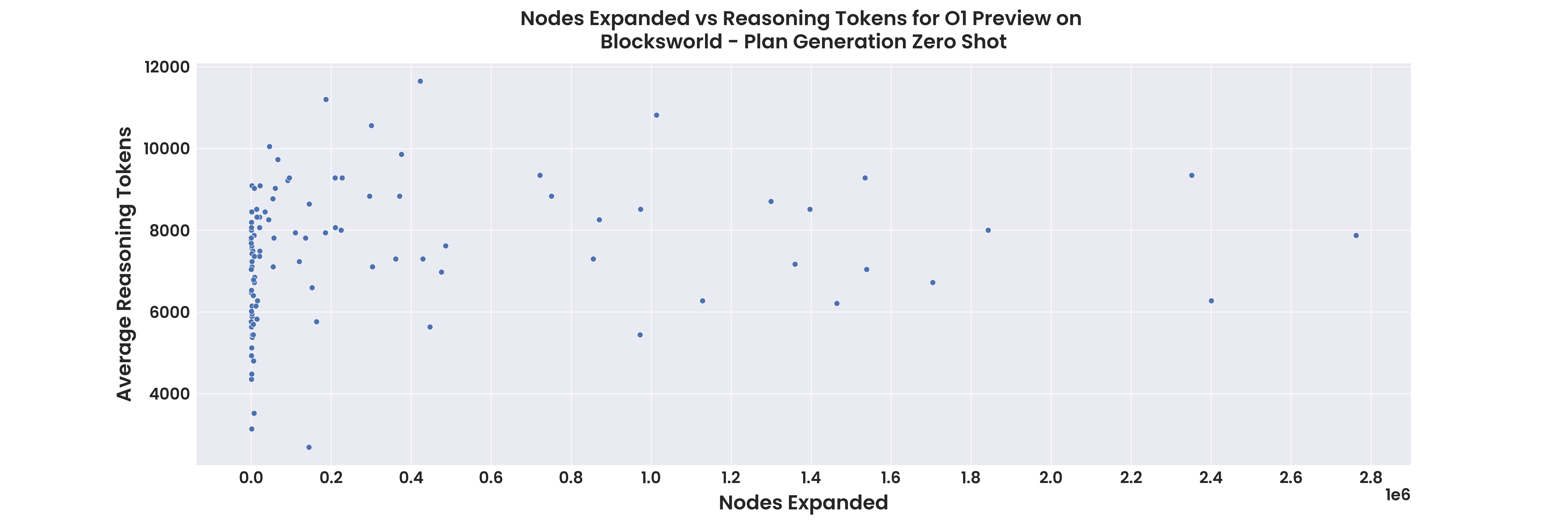}
    \caption{The number of reasoning tokens used by o1-preview when solving Blocksworld instances does not track the number of nodes that need to be expanded to solve the problem.}
    \label{fig:nodes_tokens}
\end{figure}
\subsection{Prompt to Translate From Mystery Back to Blocksworld}
\label{a:mystery_mapping}
\begin{lstlisting}
I am playing with a set of objects. Here are the actions I can do

      Attack object
      Feast object from another object
      Succumb object
      Overcome object from another object
      
I have the following restrictions on my actions:
To perform Attack action, the following facts need to be true: Province object, Planet object, Harmony.
Once Attack action is performed the following facts will be true: Pain object.
Once Attack action is performed the following facts will be false: Province object, Planet object, Harmony.
To perform Succumb action, the following facts need to be true: Pain object.
Once Succumb action is performed the following facts will be true: Province object, Planet object, Harmony.
Once Succumb action is performed the following facts will be false: Pain object.
To perform Overcome action, the following needs to be true: Province other object, Pain object.
Once Overcome action is performed the following will be true: Harmony, Province object, Object Craves other object.
Once Overcome action is performed the following will be false: Province other object, Pain object.
To perform Feast action, the following needs to be true: Object Craves other object, Province object, Harmony.
Once Feast action is performed the following will be true: Pain object, Province other object.
Once Feast action is performed the following will be false:, Object Craves other object, Province object, Harmony.


You will be given a set of initial conditions and a goal condition. To solve the problem, you will have to tell me which actions to take and in which order in order to achieve the goal.

Please provide your answers using the above terminology. However, you may find it helpful to translate the above description into a common-sense format while working out your solution. Just remember to translate it back later!
Instead of thinking in terms of "objects", think in terms of different alphabet blocks (block A, block B, etc.) which you are stacking (using just one hand) in towers on a table.

Then the "facts" that are true or false at a given time are really facts about the blocks and the hand:
      "Province object a" just means that "block A is clear" or, equivalently, "nothing is on top of block A"
      "Planet object a" is another way of saying "block A is on the table"
      "Harmony" is a codeword for "my hand isn't holding anything"
      "Pain object a" = "the hand is holding block A"
      "object a Craves object b" translates to "block A is on top of block B"

And the "actions" can be seen as stacking and unstacking of blocks (where the restrictions stop us from picking up the bottom block in a tower or holding more than one block in the hand at a time):
      "Attack object a" translates to "pick up block A directly from the table"
      "Feast object a from object b" translates to "pick up block A from directly on top of block B"
      "Succumb object a" translates to "put block A directly on the table"
      "Overcome object a from object b" translates to "put block A directly on top of block B"
      
If you do use this framing, make sure to remember to translate back into the original terms.
\end{lstlisting}
\subsection{Prompts for Blocksworld}
\subsubsection{(Solvable) Blocksworld Instances - Zero-Shot in Natural Language}
\begin{lstlisting}
I am playing with a set of blocks where I need to arrange the blocks into stacks. Here are the actions I can do

Pick up a block
Unstack a block from on top of another block
Put down a block
Stack a block on top of another block

I have the following restrictions on my actions:
I can only pick up or unstack one block at a time.
I can only pick up or unstack a block if my hand is empty.
I can only pick up a block if the block is on the table and the block is clear. A block is clear if the block has no other blocks on top of it and if the block is not picked up.
I can only unstack a block from on top of another block if the block I am unstacking was really on top of the other block.
I can only unstack a block from on top of another block if the block I am unstacking is clear.
Once I pick up or unstack a block, I am holding the block.
I can only put down a block that I am holding.
I can only stack a block on top of another block if I am holding the block being stacked.
I can only stack a block on top of another block if the block onto which I am stacking the block is clear.
Once I put down or stack a block, my hand becomes empty.
Once you stack a block on top of a second block, the second block is no longer clear.

[STATEMENT]
As initial conditions I have that, the red block is clear, the blue block is clear, the yellow block is clear, the hand is empty, the blue block is on top of the orange block, the red block is on the table, the orange block is on the table and the yellow block is on the table.
My goal is to have that the orange block is on top of the blue block.

What is the plan to achieve my goal? Just give the actions in the plan.


=============o1-preview RESPONSE================
To achieve your goal of stacking the orange block on top of the blue block, you can follow these steps:

1. **Unstack the blue block from on top of the orange block.**
   - Since the blue block is clear and your hand is empty, you can unstack it from the orange block.
2. **Put down the blue block on the table.**
   - With the blue block in your hand, you can place it onto the table.
3. **Pick up the orange block from the table.**
   - The orange block is now clear (no blocks on top of it), and your hand is empty after putting down the blue block, so you can pick it up.
4. **Stack the orange block on top of the blue block.**
   - The blue block is clear on the table, so you can stack the orange block you're holding onto it.

**Actions in the plan:**

- **Unstack** the blue block from the orange block.
- **Put down** the blue block on the table.
- **Pick up** the orange block from the table.
- **Stack** the orange block on top of the blue block.\end{lstlisting}

\subsubsection{(Solvable) Harder Blocksworld Instances - Zero-Shot in PDDL}
\begin{lstlisting}
The following is a PDDL specification of a planning problem. The first part, under the heading [DOMAIN], is the domain file. The second part, under the heading [QUERY PROBLEM], is the problem file. Using this information, which is correct, and no further assumptions, find a plan which, when run from the specified initial state, satisfies the specified goal. Provide your answer as a sequence of actions in PDDL format. An action ACTION which acts on two objects OBJ1 and OBJ2 would be written (ACTION OBJ1 OBJ2). Do not provide anything else in your answer.
[DOMAIN]
(define (domain blocksworld-4ops)
  (:requirements :strips)
(:predicates (clear ?x)
             (ontable ?x)
             (handempty)
             (holding ?x)
             (on ?x ?y))

(:action pick-up
  :parameters (?ob)
  :precondition (and (clear ?ob) (ontable ?ob) (handempty))
  :effect (and (holding ?ob) (not (clear ?ob)) (not (ontable ?ob))
               (not (handempty))))

(:action put-down
  :parameters  (?ob)
  :precondition (holding ?ob)
  :effect (and (clear ?ob) (handempty) (ontable ?ob)
               (not (holding ?ob))))

(:action stack
  :parameters  (?ob ?underob)
  :precondition (and (clear ?underob) (holding ?ob))
  :effect (and (handempty) (clear ?ob) (on ?ob ?underob)
               (not (clear ?underob)) (not (holding ?ob))))

(:action unstack
  :parameters  (?ob ?underob)
  :precondition (and (on ?ob ?underob) (clear ?ob) (handempty))
  :effect (and (holding ?ob) (clear ?underob)
               (not (on ?ob ?underob)) (not (clear ?ob)) (not (handempty)))))

[QUERY PROBLEM]
(define(problem BW-rand-6)
(:domain blocksworld-4ops)
(:objects a b c d e f )
(:init
(handempty)
(ontable a)
(on b f)
(on c a)
(on d b)
(on e c)
(on f e)
(clear d)
)
(:goal
(and
(on a f)
(on b c)
(on c a)
(on d b)
(on f e))
)
)
)

[PLAN]

=============o1-preview RESPONSE================
(unstack d b)
(put-down d)
(unstack b f)
(stack b c)
(pick-up d)
(stack d b)
(unstack f e)
(put-down f)
(unstack e c)
(put-down e)
(pick-up f)
(stack f e)
(unstack c a)
(put-down c)
(pick-up a)
(stack a f)
(pick-up c)
(stack c a)
(pick-up b)
(stack b c)
(pick-up d)
(stack d b)\end{lstlisting}
\subsubsection{Unsolvable Blocksworld Instances - Zero-Shot in PDDL}
\begin{lstlisting}
The following is a PDDL specification of a planning problem. The first part, under the heading [DOMAIN], is the domain file. The second part, under the heading [QUERY PROBLEM], is the problem file. Using this information, which is correct, and no further assumptions, find a plan which, when run from the specified initial state, satisfies the specified goal. Provide your answer as a sequence of actions in PDDL format. An action ACTION which acts on two objects OBJ1 and OBJ2 would be written (ACTION OBJ1 OBJ2). Do not provide anything else in your answer.
[DOMAIN]
(define (domain blocksworld-4ops)
  (:requirements :strips)
(:predicates (clear ?x)
             (ontable ?x)
             (handempty)
             (holding ?x)
             (on ?x ?y))

(:action pick-up
  :parameters (?ob)
  :precondition (and (clear ?ob) (ontable ?ob) (handempty))
  :effect (and (holding ?ob) (not (clear ?ob)) (not (ontable ?ob))
               (not (handempty))))

(:action put-down
  :parameters  (?ob)
  :precondition (holding ?ob)
  :effect (and (clear ?ob) (handempty) (ontable ?ob)
               (not (holding ?ob))))

(:action stack
  :parameters  (?ob ?underob)
  :precondition (and (clear ?underob) (holding ?ob))
  :effect (and (handempty) (clear ?ob) (on ?ob ?underob)
               (not (clear ?underob)) (not (holding ?ob))))

(:action unstack
  :parameters  (?ob ?underob)
  :precondition (and (on ?ob ?underob) (clear ?ob) (handempty))
  :effect (and (holding ?ob) (clear ?underob)
               (not (on ?ob ?underob)) (not (clear ?ob)) (not (handempty)))))

[QUERY PROBLEM]
(define(problem BW-rand-4)
(:domain blocksworld-4ops)
(:objects a b c d )
(:init
(handempty)
(on a b)
(on b c)
(on c d)
(ontable d)
(clear a)
)
(:goal
        (and
                (on d c)
                (on c d)
        )
)
)

[PLAN]
=============o1-preview RESPONSE================
**No plan exists that satisfies the goal with the given initial state and domain.**\end{lstlisting}
\subsection{Prompts for Mystery Blocksworld}
\subsubsection{Mystery Blocksworld Instances - Zero-Shot in Natural Language}
\begin{lstlisting}
I am playing with a set of objects. Here are the actions I can do

   Attack object
   Feast object from another object
   Succumb object
   Overcome object from another object

I have the following restrictions on my actions:
    To perform Attack action, the following facts need to be true: Province object, Planet object, Harmony.
    Once Attack action is performed the following facts will be true: Pain object.
    Once Attack action is performed the following facts will be false: Province object, Planet object, Harmony.
    To perform Succumb action, the following facts need to be true: Pain object.
    Once Succumb action is performed the following facts will be true: Province object, Planet object, Harmony.    
    Once Succumb action is performed the following facts will be false: Pain object.
    To perform Overcome action, the following needs to be true: Province other object, Pain object.
    Once Overcome action is performed the following will be true: Harmony, Province object, Object Craves other object.
    Once Overcome action is performed the following will be false: Province other object, Pain object.
    To perform Feast action, the following needs to be true: Object Craves other object, Province object, Harmony.
    Once Feast action is performed the following will be true: Pain object, Province other object.
    Once Feast action is performed the following will be false:, Object Craves other object, Province object, Harmony.

[STATEMENT]
As initial conditions I have that, object b craves object c, harmony, planet object a, planet object c, planet object d, province object a, province object b and province object d.
My goal is to have that object c craves object b.

What is the plan to achieve my goal? Just give the actions in the plan.
=============o1-preview RESPONSE================
To achieve the goal where object c craves object b, you can perform the following sequence of actions:

1. **Feast object b from object c**
   - Preconditions:
     - `Object b craves object c` (initial fact)
     - `Province object b` (initial fact)
     - `Harmony` (initial fact)
   - Effects:
     - Adds: `Pain object b`, `Province object c`
     - Deletes: `Object b craves object c`, `Province object b`, `Harmony`

2. **Succumb object b**
   - Preconditions:
     - `Pain object b` (from previous action)
   - Effects:
     - Adds: `Province object b`, `Planet object b`, `Harmony`
     - Deletes: `Pain object b`

3. **Attack object c**
   - Preconditions:
     - `Province object c` (from step 1)
     - `Planet object c` (initial fact)
     - `Harmony` (restored in step 2)
   - Effects:
     - Adds: `Pain object c`
     - Deletes: `Province object c`, `Planet object c`, `Harmony`

4. **Overcome object c from object b**
   - Preconditions:
     - `Province object b` (restored in step 2)
     - `Pain object c` (from step 3)
   - Effects:
     - Adds: `Harmony`, `Province object c`, `Object c craves object b` (goal achieved)
     - Deletes: `Province object b`, `Pain object c`

By following this plan, you manipulate the states of the objects through the actions provided, ultimately achieving the desired state where `Object c craves object b`.

**Answer:**
To achieve the goal, perform these actions in order:

1. Feast object b from object c
2. Succumb object b
3. Attack object c
4. Overcome object c from object b\end{lstlisting}

\subsection{Prompts for Randomized Mystery Blocksworld}
\subsubsection{Randomized Mystery Blocksworld Instances - Zero-Shot in Natural Language}
\begin{lstlisting}

I am playing with a set of objects. Here are the actions I can do

J4gv801gnu2it0yj object_0 object_1.
U64y1a9apusmslxb object_0.
Kip9uw781pv62umn object_0.
Wio5amhq7814n006 object_0 object_1.

I have the following restrictions on my actions:
To perform j4gv801gnu2it0yj action, the following facts need to be true: tv30k33pzoulql6w object_1, and a46zhwykn6jvbabk object_0
Once j4gv801gnu2it0yj is performed the following facts will be true: cql4o62p1yeke3ok, tv30k33pzoulql6w object_0, and b6e9q4r60gagvdcn object_0 object_1
Once j4gv801gnu2it0yj is performed the following facts will be false: tv30k33pzoulql6w object_1, and a46zhwykn6jvbabk object_0
To perform u64y1a9apusmslxb action, the following facts need to be true: tv30k33pzoulql6w object_0, af9tse23ljclsqad object_0, and cql4o62p1yeke3ok
Once u64y1a9apusmslxb is performed the following facts will be true: a46zhwykn6jvbabk object_0
Once u64y1a9apusmslxb is performed the following facts will be false: tv30k33pzoulql6w object_0, af9tse23ljclsqad object_0, and cql4o62p1yeke3ok
To perform kip9uw781pv62umn action, the following facts need to be true:a46zhwykn6jvbabk object_0
Once kip9uw781pv62umn is performed the following facts will be true: tv30k33pzoulql6w object_0, cql4o62p1yeke3ok, and af9tse23ljclsqad object_0
Once kip9uw781pv62umn is performed the following facts will be false: a46zhwykn6jvbabk object_0
To perform wio5amhq7814n006 action, the following facts need to be true: b6e9q4r60gagvdcn object_0 object_1, tv30k33pzoulql6w object_0, and cql4o62p1yeke3ok
Once wio5amhq7814n006 is performed the following facts will be true: a46zhwykn6jvbabk object_0, and tv30k33pzoulql6w object_1
Once wio5amhq7814n006 is performed the following facts will be false: b6e9q4r60gagvdcn object_0 object_1, tv30k33pzoulql6w object_0, and cql4o62p1yeke3ok

[STATEMENT]
As initial conditions I have that, af9tse23ljclsqad object_0, af9tse23ljclsqad object_2, af9tse23ljclsqad object_3, b6e9q4r60gagvdcn object_1 object_2, cql4o62p1yeke3ok, tv30k33pzoulql6w object_0, tv30k33pzoulql6w object_1 and tv30k33pzoulql6w object_3.
My goal is to have that b6e9q4r60gagvdcn object_2 object_1.

What is the plan to achieve my goal? Just give the actions in the plan.
=============o1-preview RESPONSE================
To achieve the goal `b6e9q4r60gagvdcn object_2 object_1`, you need to perform the following sequence of actions:

1. **wio5amhq7814n006 object_1 object_2**
   - This action reverses the relationship between `object_1` and `object_2`, setting up the necessary conditions for the next steps.
2. **kip9uw781pv62umn object_1**
   - By performing this action on `object_1`, you restore certain facts (`cql4o62p1yeke3ok`) required for subsequent actions and adjust the state of `object_1`.
3. **u64y1a9apusmslxb object_2**
   - This action modifies `object_2` to have the required property `a46zhwykn6jvbabk object_2`, which is a precondition for the final action.
4. **j4gv801gnu2it0yj object_2 object_1**
   - Performing this action with `object_2` and `object_1` establishes the goal fact `b6e9q4r60gagvdcn object_2 object_1`.

**Plan Summary:**

- **Action 1:** `wio5amhq7814n006 object_1 object_2`
- **Action 2:** `kip9uw781pv62umn object_1`
- **Action 3:** `u64y1a9apusmslxb object_2`
- **Action 4:** `j4gv801gnu2it0yj object_2 object_1`

This sequence of actions satisfies all the necessary preconditions step-by-step and ultimately achieves your goal.
\end{lstlisting}
\subsubsection{Unsolvable Randomized Mystery Blocksworld Instances - Zero-Shot in PDDL}
\begin{lstlisting}
The following is a PDDL specification of a planning problem. The first part, under the heading [DOMAIN], is the domain file. The second part, under the heading [QUERY PROBLEM], is the problem file. Using this information, which is correct, and no further assumptions, find a plan which, when run from the specified initial state, satisfies the specified goal. Provide your answer as a sequence of actions in PDDL format. An action ACTION which acts on two objects OBJ1 and OBJ2 would be written (ACTION OBJ1 OBJ2). Do not provide anything else in your answer.
[DOMAIN]
(define (domain xaji0y)
    (:requirements :equality)
    (:predicates
        (tv30k33pzoulql6w ?x1 - object)
        (af9tse23ljclsqad ?x1 - object)
        (cql4o62p1yeke3ok )
        (a46zhwykn6jvbabk ?x1 - object)
        (b6e9q4r60gagvdcn ?x1 - object ?x2 - object)
    )
    
    (:action u64y1a9apusmslxb
     :parameters (?ob1 - object)
     :precondition (and (tv30k33pzoulql6w ?ob1) (af9tse23ljclsqad ?ob1) (cql4o62p1yeke3ok ))
     :effect (and
        (a46zhwykn6jvbabk ?ob1)
        (not (tv30k33pzoulql6w ?ob1))
        (not (af9tse23ljclsqad ?ob1))
        (not (cql4o62p1yeke3ok )))
    )


    (:action kip9uw781pv62umn
     :parameters (?ob1 - object)
     :precondition (a46zhwykn6jvbabk ?ob1)
     :effect (and
        (tv30k33pzoulql6w ?ob1)
        (cql4o62p1yeke3ok )
        (af9tse23ljclsqad ?ob1)
        (not (a46zhwykn6jvbabk ?ob1)))
    )


    (:action j4gv801gnu2it0yj
     :parameters (?ob1 - object ?ob2 - object)
     :precondition (and (tv30k33pzoulql6w ?ob2) (a46zhwykn6jvbabk ?ob1))
     :effect (and
        (cql4o62p1yeke3ok )
        (tv30k33pzoulql6w ?ob1)
        (b6e9q4r60gagvdcn ?ob1 ?ob2)
        (not (tv30k33pzoulql6w ?ob2))
        (not (a46zhwykn6jvbabk ?ob1)))
    )


    (:action wio5amhq7814n006
     :parameters (?ob1 - object ?ob2 - object)
     :precondition (and (b6e9q4r60gagvdcn ?ob1 ?ob2) (tv30k33pzoulql6w ?ob1) (cql4o62p1yeke3ok ))
     :effect (and
        (a46zhwykn6jvbabk ?ob1)
        (tv30k33pzoulql6w ?ob2)
        (not (b6e9q4r60gagvdcn ?ob1 ?ob2))
        (not (tv30k33pzoulql6w ?ob1))
        (not (cql4o62p1yeke3ok )))
    )

)

[QUERY PROBLEM]
(define(problem instance-1-4)
(:domain xaji0y)
(:objects a b c d )
(:init
(cql4o62p1yeke3ok)
(b6e9q4r60gagvdcn a d)
(af9tse23ljclsqad b)
(b6e9q4r60gagvdcn c a)
(b6e9q4r60gagvdcn d b)
(tv30k33pzoulql6w c)
)
(:goal
        (and
                (b6e9q4r60gagvdcn a d)
                (b6e9q4r60gagvdcn b a)
                (b6e9q4r60gagvdcn b d)
        )
)
)

[PLAN]
=============o1-preview RESPONSE================
There is no sequence of actions that, when run from the specified initial state, satisfies the specified goal. Thus, no plan can be provided.
\end{lstlisting}

\end{document}